\documentclass{article}
\usepackage{spconf,amsmath,graphicx}

\usepackage{enumitem}
\setlist{nosep, leftmargin=14pt}

\usepackage{mwe} 

\usepackage{multirow}
\usepackage{booktabs}
\usepackage{tikz}

\title{Contrastive Language Prompting to Ease \\False Positives in Medical Anomaly Detection}

\name{YeongHyeon Park\quad{} Myung Jin Kim\quad{} Hyeong Seok Kim}
\address{ 
    SK Planet Co., Ltd. \\
    \small{\texttt{\{yeonghyeon, myungjin, beman\}@sk.com}}
}

\begin{document}

\maketitle

\begin{abstract}
A pre-trained visual-language model, contrastive language-image pre-training (CLIP), successfully accomplishes various downstream tasks with text prompts, such as finding images or localizing regions within the image.
Despite CLIP’s strong multi-modal data capabilities, it remains limited in specialized environments, such as medical applications.
For this purpose, many CLIP variants\textemdash{}i.e., BioMedCLIP, and MedCLIP-SAMv2\textemdash{}have emerged, but false positives related to normal regions persist.
Thus, we aim to present a simple yet important goal of reducing false positives in medical anomaly detection.
We introduce a \textit{\textbf{C}ontrastive \textbf{LA}nguage \textbf{P}rompting (\textbf{CLAP})} method that leverages both positive and negative text prompts.
This straightforward approach identifies potential lesion regions by visual attention to the positive prompts in the given image. 
To reduce false positives, we attenuate attention on normal regions using negative prompts.
Extensive experiments with the BMAD dataset, including six biomedical benchmarks, demonstrate that CLAP method enhances anomaly detection performance.
Our future plans include developing an automated fine prompting method for more practical usage.
\end{abstract}
\section{Introduction}
\label{sec:intro}

\begin{figure}[t]
    \scriptsize
    \setlength{\tabcolsep}{0pt}
    \centering
    \resizebox{\columnwidth}{!}{%
        \begin{tabular}{c}
            \begin{tabular}{cc ccccc}
                \vspace{-0.4cm} \\
                \qquad{} & \qquad{} & Input & Ground-truth & $A_{\textit{positive}}$ & $A_{\textit{negative}}$ & $A_{\textit{CLAP}}$ (Ours) \\
                
                \multirow{2}{*}{\rotatebox[origin=c]{90}{\textbf{Normal}}} & 
                \rotatebox[origin=c]{90}{\qquad{}\qquad{}\qquad{}\quad{}Brain MRI} &
                \includegraphics[width=0.201\columnwidth]{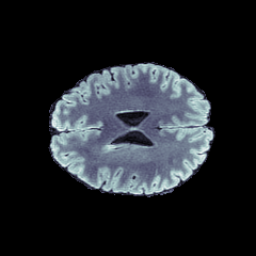} &
                \includegraphics[width=0.201\columnwidth]{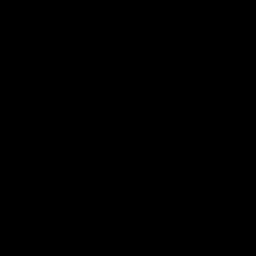} &
                \begin{tikzpicture}
                    \node[anchor=south west,inner sep=0] (image) at (0,0) {\includegraphics[width=0.201\columnwidth]{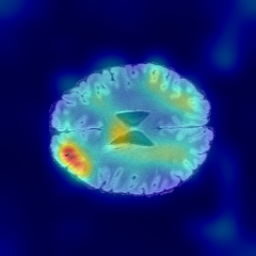}};
                    \begin{scope}[x={(image.south east)},y={(image.north west)}]
                        \draw[red, thick, ->] (0.55, 0.78) -- (0.7, 0.6); 
                        \draw[red, thick, ->] (0.55, 0.78) -- (0.43, 0.5); 
                        \node[red] at (0.5, 0.85) {False positives}; 
                    \end{scope}
                \end{tikzpicture} &
                \begin{tikzpicture}
                    \node[anchor=south west,inner sep=0] (image) at (0,0) {\includegraphics[width=0.201\columnwidth]{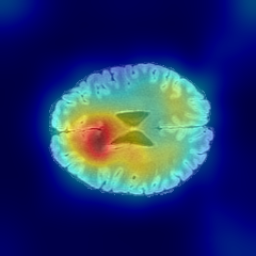}};
                    \begin{scope}[x={(image.south east)},y={(image.north west)}]
                        \draw[green, thick, ->] (0.55, 0.78) -- (0.7, 0.6); 
                        \draw[green, thick, ->] (0.55, 0.78) -- (0.43, 0.5); 
                        \node[green] at (0.5, 0.85) {True negatives}; 
                    \end{scope}
                \end{tikzpicture} &
                \includegraphics[width=0.201\columnwidth]{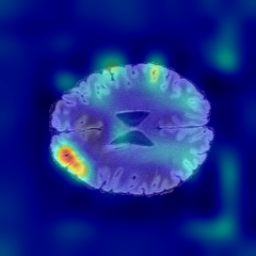} \\
                \vspace{-1.60cm} \\

                &
                \rotatebox[origin=c]{90}{\qquad{}\qquad{}\qquad{}\quad{}Liver CT} &
                \includegraphics[width=0.201\columnwidth]{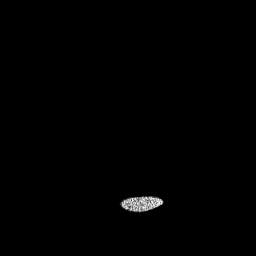} &
                \includegraphics[width=0.201\columnwidth]{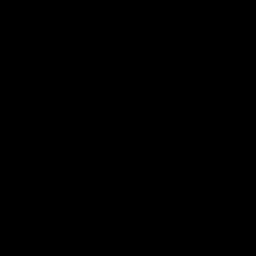} &
                \begin{tikzpicture}
                    \node[anchor=south west,inner sep=0] (image) at (0,0) {\includegraphics[width=0.201\columnwidth]{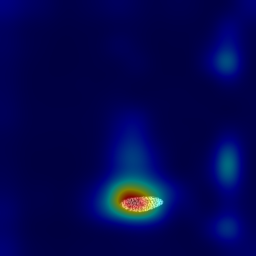}};
                    \begin{scope}[x={(image.south east)},y={(image.north west)}]
                        \draw[red, thick, ->] (0.45, 0.75) -- (0.5, 0.28); 
                        \node[red] at (0.5, 0.85) {False positives}; 
                    \end{scope}
                \end{tikzpicture} &
                \begin{tikzpicture}
                    \node[anchor=south west,inner sep=0] (image) at (0,0) {\includegraphics[width=0.201\columnwidth]{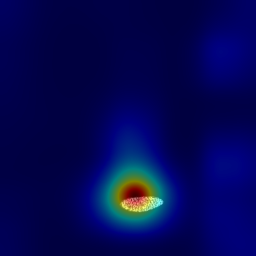}};
                    \begin{scope}[x={(image.south east)},y={(image.north west)}]
                        \draw[green, thick, ->] (0.45, 0.75) -- (0.5, 0.28); 
                        \node[green] at (0.5, 0.85) {True negatives}; 
                    \end{scope}
                \end{tikzpicture} &
                \includegraphics[width=0.201\columnwidth]{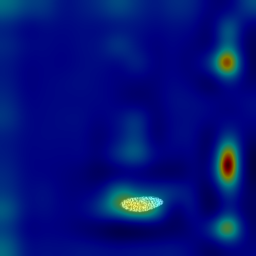} \\
                \vspace{-1.61cm} \\
                
                \hline
                \vspace{-0.2cm} \\
                
                \multirow{2}{*}{\rotatebox[origin=c]{90}{\textbf{Abnormal}}} & 
                \rotatebox[origin=c]{90}{\qquad{}\qquad{}\qquad{}\quad{}Brain MRI} &
                \includegraphics[width=0.201\columnwidth]{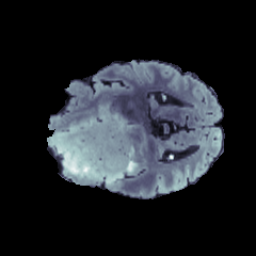} &
                \includegraphics[width=0.201\columnwidth]{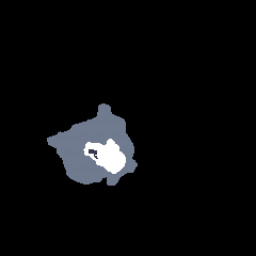} &
                \begin{tikzpicture}
                    \node[anchor=south west,inner sep=0] (image) at (0,0) {\includegraphics[width=0.201\columnwidth]{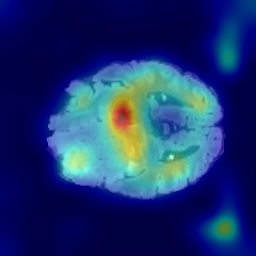}};
                    \begin{scope}[x={(image.south east)},y={(image.north west)}]
                        \draw[red, thick, ->] (0.55, 0.22) -- (0.7, 0.6); 
                        \node[red] at (0.5, 0.15) {False positives}; 
                    \end{scope}
                \end{tikzpicture} &
                \begin{tikzpicture}
                    \node[anchor=south west,inner sep=0] (image) at (0,0) {\includegraphics[width=0.201\columnwidth]{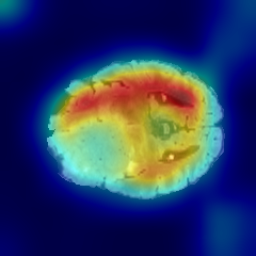}};
                    \begin{scope}[x={(image.south east)},y={(image.north west)}]
                        \draw[green, thick, ->] (0.55, 0.22) -- (0.7, 0.6); 
                        \node[green] at (0.5, 0.15) {True negatives}; 
                    \end{scope}
                \end{tikzpicture} &
                \includegraphics[width=0.201\columnwidth]{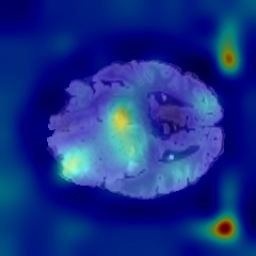} \\
                \vspace{-1.60cm} \\

                &
                \rotatebox[origin=c]{90}{\qquad{}\qquad{}\qquad{}\quad{}Liver CT} &
                \includegraphics[width=0.201\columnwidth]{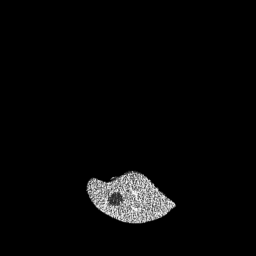} &
                \includegraphics[width=0.201\columnwidth]{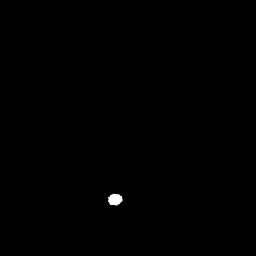} &
                \begin{tikzpicture}
                    \node[anchor=south west,inner sep=0] (image) at (0,0) {\includegraphics[width=0.201\columnwidth]{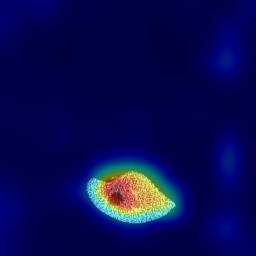}};
                    \begin{scope}[x={(image.south east)},y={(image.north west)}]
                        \draw[red, thick, ->] (0.45, 0.75) -- (0.52, 0.28); 
                        \node[red] at (0.5, 0.85) {False positives}; 
                    \end{scope}
                \end{tikzpicture} &
                \begin{tikzpicture}
                    \node[anchor=south west,inner sep=0] (image) at (0,0) {\includegraphics[width=0.201\columnwidth]{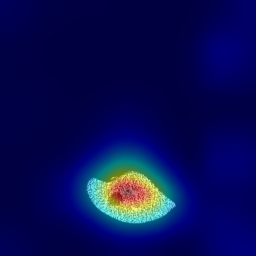}};
                    \begin{scope}[x={(image.south east)},y={(image.north west)}]
                        \draw[green, thick, ->] (0.45, 0.75) -- (0.52, 0.28); 
                        \node[green] at (0.5, 0.85) {True negatives}; 
                    \end{scope}
                \end{tikzpicture} &
                \includegraphics[width=0.201\columnwidth]{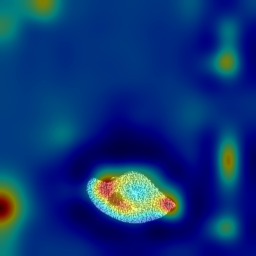} \\
                \vspace{-1.67cm} \\

            \end{tabular}
        \end{tabular}
    }
    \caption{Generated attention maps by leveraging BiomedCLIP~\cite{zhang2023biomedclip}. $A_{\textit{positive}}$ and $A_{\textit{negative}}$ are the attention maps obtained using positive or negative prompts only. $A_{\textit{CLAP}}$ shows results of our proposal, dubbed \textit{\textbf{C}ontrastive \textbf{LA}nguage \textbf{P}rompting (\textbf{CLAP})}. CLAP leverages both positive and negative prompts. The negative prompts are used to attenuate false positive attention of normal regions.}
    \label{fig:overview}
    \vspace{-0.4cm}
\end{figure}
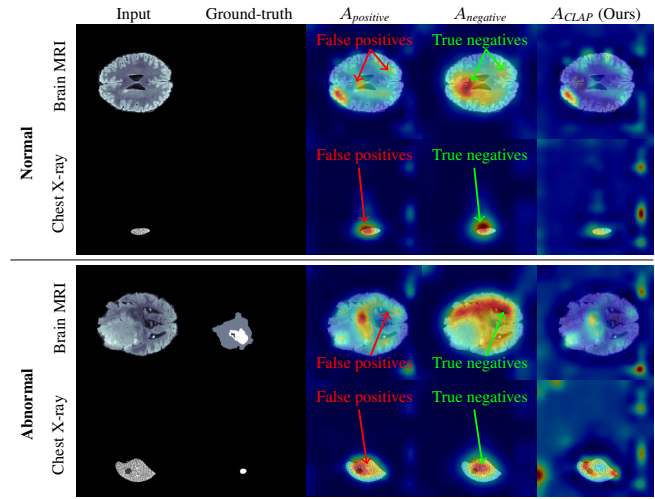

\begin{figure*}[t]
    \resizebox{\linewidth}{!}{%
        \includegraphics[width=1.0\linewidth,trim={0.0cm 0.0cm 0.0cm 0.0cm},clip]{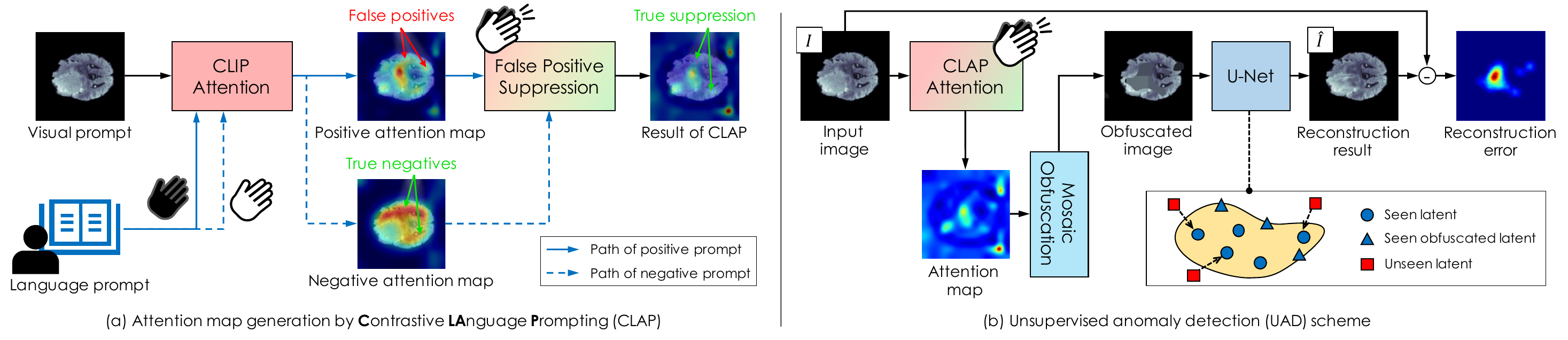}
    }
    \vspace{-0.7cm}
    \caption{Schematic diagram of our method. Existing positive prompt methods only utilize positive prompts. In this situation, the false positive attention issue remains. In comparison, our method CLAP successfully suppresses false positives by additionally exploiting negative prompts, shown in (a). After getting the attention map of CLAP, we employ the existing UAD model EAR~\cite{park2024visual}. We only replace the saliency map for mosaic obfuscation with an attention map from CLAP, shown in (b).}
    \label{fig:cpmethod}
    \vspace{-0.2cm}
\end{figure*}

Recent advancements in multi-modal models, particularly visual-language models (VLMs), have revolutionized various downstream tasks, such as image retrieval, captioning, and object localization. Among these, Contrastive Language–Image Pre-training (CLIP)~\cite{radford2021learning} has demonstrated remarkable performance by leveraging natural language prompts to interpret visual data. This capability enables CLIP to handle a wide array of tasks without specialized fine-tuning. However, its application to highly specialized domains, such as medical imaging, has uncovered limitations.

In the medical field, accurate anomaly detection is crucial for early diagnosis and treatment. Nevertheless, general-purpose models like CLIP often struggle with the intricacies of medical images, which contain subtle and features unique to medical imaging essential for identifying pathological regions. To improve performance in biomedical domains, various adaptations of CLIP, such as BiomedCLIP~\cite{zhang2023biomedclip} and MedCLIP-SAMv2~\cite{koleilat2024medclip}, have been proposed to improve performance in biomedical domains. They shows surprising enhancement of medical reasoning compared to ordinary models, but the issue of false positives\textemdash{}incorrectly identifying normal regions as anomalous\textemdash{}remains prevalent. These false positives can lead to unnecessary medical procedures, increasing the burden on healthcare systems and potentially harming patients.

To address this issue, we propose a novel method called \textit{\textbf{C}ontrastive \textbf{LA}nguage \textbf{P}rompting (\textbf{CLAP})}, which introduces a more refined way of leveraging natural language prompts for medical anomaly detection. By leveraging both positive and negative prompts, our method aims to find out lesions accurately with CLIP attention. Positive prompts guide the CLIP attention toward potential lesion regions, while negative prompts help attenuate the attention on normal regions, thereby reducing the occurrence of strong attention to false positives. This approach not only provides a more improved understanding of the medical image but also aligns with the demands for reliability in medical diagnostics by artificial intelligence.

We can just determine whether disease or not based on the CLIP attention. Toward a more accurate diagnosis, we employ an unsupervised anomaly detection (UAD) method that features a reconstruction-by-inpainting strategy~\cite{park2024visual}. For this, we obfuscate strong attention regions, over $\mu+0.674\sigma$ ($Q3$)~\cite{tukey1977exploratory} valued regions, by considering suspected disease regions. Then, we attempt to reconstruct obfuscated regions into normal patterns by U-Net which is trained with normal samples only. Finally, we determine the final disease based on the reconstruction error obtained.

To evaluate the legitimacy of our proposal, we perform extensive experiments using BMAD dataset~\cite{bao2024bmad}. This dataset provides six benchmarks for five anatomies. Visual comparisons demonstrate that CLAP successfully overcomes issues of strong attention in non-lesion regions. In addition, we improved UAD performance compared to existing methods. Through this work, we aim to bridge the gap between general-purpose VLMs and the specific needs of medical anomaly detection. We conclude by discussing the potential for automating language prompt construction to further improve the usability of this approach in real-world clinical settings.

\section{Methods}
\label{sec:method}
In this section, we introduce a VLM-leveraged medical UAD method. The proposed method as shown in Fig.~\ref{fig:cpmethod} consists of two major components as follows. (1) Obtain an attention map that indicates a suspected lesion area by zero-shot inference of VLM. For this, we propose a method that overcomes an issue of false positive attention in a single positive language prompting. (2) We present a UAD method using a reconstruction-by-inpainting strategy. The UAD model attempts to reconstruct a suspected lesion area covered by an attention mask. Then, it can be determined whether lesion or not based on the reconstruction error. In-depth details are introduced sequentially.

\subsection{Contrastive language prompting method}
\label{subsec:cp}
The VLM takes each image $I$ as a visual prompt and each text string $T$ as a language prompt. All CLIP variants are consisted with image encoder $\Phi_{I}$ and text encoder $\Phi_{T}$. In this study, we exploit BiomedCLIP~\cite{zhang2023biomedclip} which has special feature extraction capability on the biomedical domain by the fine-tuning process. Specifically, we simply inherit an attention map generation method of MedCLIP-SAMv2~\cite{koleilat2024medclip} summarized as \eqref{eq:saliencymap}. Note, $MI$ is the mutual information operation and $\beta$ is a hyperparameter to balance each term.

\begin{equation}
    \begin{aligned}
        A = MI(Z_{I}, Z_{T}; \theta{}_{I}, \theta{}_{T}) - \beta{} \times MI(Z_{I}, I; \theta{}_{I}, \theta{}_{T}) \\
        w.r.t.\ \ Z_{I} = \Phi_{I}(I; \theta{}_{I}), Z_{T} = \Phi_{T}(T; \theta{}_{T})
    \end{aligned}
    \label{eq:saliencymap}
\end{equation}

The results of a positive language prompt, denoted as $A_{\textit{positive}}$, are shown in Fig.~\ref{fig:overview}. This prompt conveys comma-separated words, recommended words by ChatGPT, that can be used to refer to lesions (e.g., ``Glioma, Astrocytoma, Oligodendroglioma \ldots"). There are many strong attention regions, dubbed false positives, in $A_{\textit{positive}}$ even if the input image was normal. To mitigate the false positive issue on $A_{\textit{positive}}$, we additionally check the results of negative prompts as shown in $A_{\textit{negative}}$. Those results show not very strong attention on non-negative regions (false negatives). Moreover, the true negative regions are mostly highlighted that can be used to suppress the false positives of $A_{\textit{positive}}$.

In this study, we propose CLAP, a straightforward approach. CLAP shows a very intuitive approach to take attention map as `$ A_{\textit{CLAP}} = A_{\textit{positive}} - A_{\textit{negative}}$'. This simple method can be further refined with a parametric function and deep neural networks. The purpose of this study is to take a first step toward reducing false positive attention when utilizing VLM for the biomedical domain. Therefore, more refined methods will be dealt with in the next study.

\begin{table}[t]\centering
    \vspace{-0.3cm}
    \caption{Positive and negative language prompt examples for each anatomy. Symbols `P' and `N' represent positive and negative respectively.}
    \vspace{0.2cm}
    \centering
    \resizebox{\columnwidth}{!}{%
        \begin{tabular}{l||c|l}
        \toprule
            \textbf{Anatomy} & \textbf{P/N} & \textbf{Language prompt} \\
        \hline
        \hline
            \multirow{2}{*}{Brain MRI} 
                & P & Glioma, Astrocytoma, Oligodendroglioma \ldots \\
                & N & Normal, Healthy gray matter \ldots \\
        \hline
            \multirow{2}{*}{Liver CT} 
                & P & Malignant cells, Dysplasia, Hyperplasia \ldots \\
                & N & Normal, Healthy, Benign \ldots \\
        \hline
            \multirow{2}{*}{Retinal OCT} 
                & P & Retinal fluid, Drusen, Retinal detachment \ldots \\
                & N & Normal, Healthy, Clear \ldots \\
        \hline
            \multirow{2}{*}{Chest X-ray} 
                & P & Consolidation, Fibrosis, Atelectasis \ldots \\
                & N & Healthy, Clear fields, Normal \ldots \\
        \hline
            \multirow{2}{*}{Lymph node} 
                & P & Metastatic carcinoma, Tumor metastasis \ldots \\
                & N & Normal, Healthy tissue \ldots \\
        \bottomrule
        \end{tabular}
    }
    \label{table:prompts}
    \vspace{-0.5cm}
\end{table}

\subsection{Reconstruction-by-inpainting for UAD}
\label{subsec:uad}
UAD approaches based on the U-Net aim to block abnormal feature transmission from the encoder to the decoder through skip connections~\cite{park2024visual}. Since it is not known in advance where and how large the anomalous pattern exists in the given image, cutting out the anomalous region by masking is difficult to prevent abnormal feature transmission.

To address this, an attention map-based saliency obfuscation method was developed~\cite{park2024visual}. We adopt this method in this study, and the overall scheme is shown in Fig.~\ref{fig:cpmethod} (b). The saliency region $S$, suspected to be abnormal, is identified using our CLAP method according to \eqref{eq:saliency}, where each pixel value exceeding $\mu+0.674\sigma$ ($Q3$)~\cite{tukey1977exploratory} is flagged. Here, $\mu$ and $\sigma$ are the mean and standard deviation of $A_{\textit{CLAP}}$.

\begin{equation}
    \begin{aligned}
        S = where(A_{\textit{CLAP}} > \mu+0.674\sigma)\\
    \end{aligned}
    \label{eq:saliency}
\end{equation}

This region will be obfuscated by a mosaic process to prevent the UAD model from receiving the abnormal pattern. The U-Net is trained to reconstruct the partially obfuscated image to its original form, a process known as reconstruction-by-inpainting. Only normal samples are used for training, ensuring the model generalizes well to normal patterns while struggling with abnormal pattern reconstruction.

During inference, the U-Net attempts to reconstruct the saliency-obfuscated image. A reconstruction error map is obtained based on the error between the reconstruction result and the original input image before obfuscation. We can perform image-level diagnosis using the maximum value of reconstruction error map.

\section{Experiments}
\label{sec:experiments}

\begin{figure}[t]
    \scriptsize
    \setlength{\tabcolsep}{0pt}
    \centering
    \resizebox{\columnwidth}{!}{%
        \begin{tabular}{c}
            \begin{tabular}{c ccc c ccc}
                \vspace{-0.4cm} \\
                \qquad{} & \multicolumn{3}{c}{\textbf{Normal}} & \; & \multicolumn{3}{c}{\textbf{Abnormal}} \\
                \qquad{} & Input & $A_{\textit{DINO}}$ & $A_{\textit{CLAP}}$ (ours) & & Input & $A_{\textit{DINO}}$ & $A_{\textit{CLAP}}$ (ours) \\

                \rotatebox[origin=c]{90}{\qquad{}\qquad{}\qquad{}\quad{}Brain MRI} &
                \includegraphics[width=0.221\columnwidth]{figures/medclip_BraTS2021_slice/test/good/00000_98_input.png} &
                \begin{tikzpicture}
                    \node[anchor=south west,inner sep=0] (image) at (0,0) {\includegraphics[width=0.221\columnwidth]{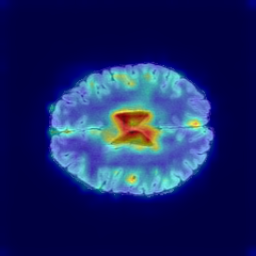}};
                    \begin{scope}[x={(image.south east)},y={(image.north west)}]
                        \draw[red, thick, ->] (0.4, 0.78) -- (0.5, 0.6); 
                        \node[red] at (0.5, 0.85) {False positives}; 
                    \end{scope}
                \end{tikzpicture} &
                \includegraphics[width=0.221\columnwidth]{figures/medclip_BraTS2021_slice/test/good/00000_98_merge_o.png} & &
                
                \includegraphics[width=0.221\columnwidth]{figures/medclip_BraTS2021_slice/test/ungood/00002_68_input.png} &
                \begin{tikzpicture}
                    \node[anchor=south west,inner sep=0] (image) at (0,0) {\includegraphics[width=0.221\columnwidth]{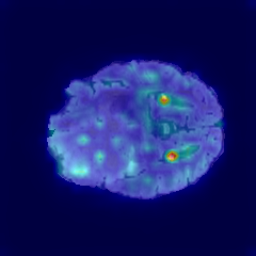}};
                    \begin{scope}[x={(image.south east)},y={(image.north west)}]
                        \draw[red, thick, ->] (0.5, 0.2) -- (0.62, 0.58); 
                        \node[red] at (0.5, 0.15) {False positives}; 
                    \end{scope}
                \end{tikzpicture} &
                \includegraphics[width=0.221\columnwidth]{figures/medclip_BraTS2021_slice/test/ungood/00002_68_merge_o.png} \\
                \vspace{-1.60cm} \\

                \rotatebox[origin=c]{90}{\qquad{}\qquad{}\qquad{}\quad{}Retinal OCT} &
                \includegraphics[width=0.221\columnwidth]{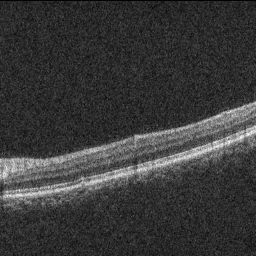} &
                \includegraphics[width=0.221\columnwidth]{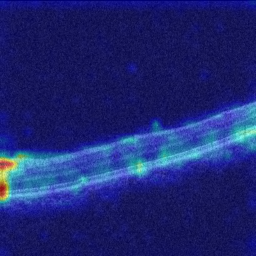} &
                \includegraphics[width=0.221\columnwidth]{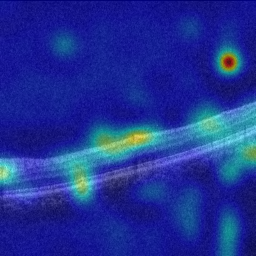} & & 
                
                \includegraphics[width=0.221\columnwidth]{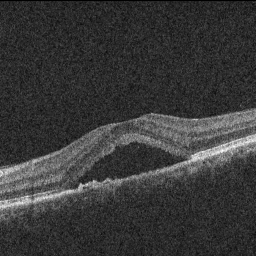} &
                \begin{tikzpicture}
                    \node[anchor=south west,inner sep=0] (image) at (0,0) {\includegraphics[width=0.221\columnwidth]{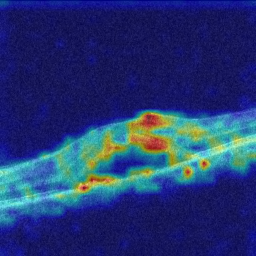}};
                    \begin{scope}[x={(image.south east)},y={(image.north west)}]
                        \draw[red, thick, ->] (0.5, 0.22) -- (0.57, 0.4); 
                        \node[red] at (0.5, 0.15) {False negatives}; 
                    \end{scope}
                \end{tikzpicture} &
                \includegraphics[width=0.221\columnwidth]{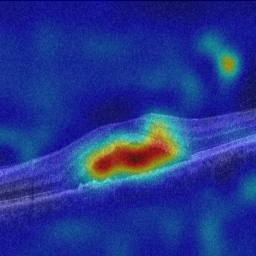} \\
                \vspace{-1.69cm} \\

                \rotatebox[origin=c]{90}{\qquad{}\qquad{}\qquad{}\quad{}Lymph node} &
                \includegraphics[width=0.221\columnwidth]{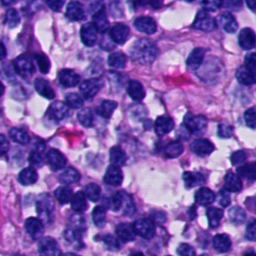} &
                \includegraphics[width=0.221\columnwidth]{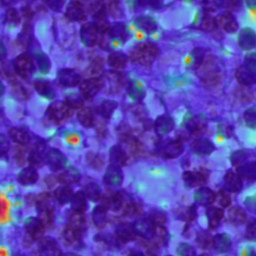} &
                \includegraphics[width=0.221\columnwidth]{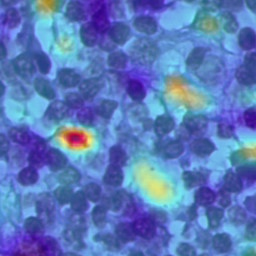} & &
                
                \includegraphics[width=0.221\columnwidth]{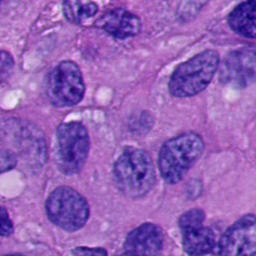} &
                \includegraphics[width=0.221\columnwidth]{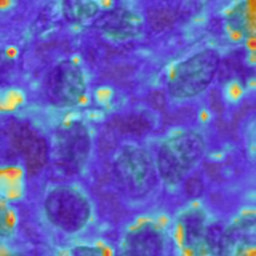} &
                \includegraphics[width=0.221\columnwidth]{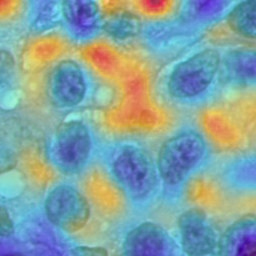} \\
                \vspace{-1.67cm} \\
            \end{tabular}
        \end{tabular}
    }
    \caption{Attention results of visual-only model and visual-language model. The visual-only model, DINO~\cite{caron2021emerging}, performs effective visual saliency attention in the ordinary domain but shows short in the medical domain. When applying our method CLAP on the visual-language model BiomedCLIP~\cite{zhang2023biomedclip}, false attentions are successfully removed.}
    \label{fig:comparison}
    \vspace{-0.4cm}
\end{figure}

\subsection{Experimental Setup}
\label{subsec:landscape}
We conduct experiments on the BMAD dataset~\cite{bao2024bmad}, which comprises brain MRI, liver CT, retinal OCT, chest X-ray, and lymph node histopathology images. The training set comprises only anomaly-free samples, whereas the test and validation sets include both anomaly-free and anomalous samples.

\noindent \textbf{Implementation Details.}
Following EAR~\cite{park2024visual}, we use a U-Net reconstruction model (Fig.~\ref{fig:cpmethod}) with five convolutional blocks in both the encoder and decoder. Each encoder feature map is directly connected to its corresponding decoder block through skip connections. The U-Net is trained to minimize reconstruction loss between the input image  $I$  and the reconstruction output $\hat{I}$. We train U-Net with 20 epochs. In each epoch, we randomly select 3k samples of the training set to shorten the experimental period at this time. We plan to conduct a full experiment in which the entire sample is trained for longer epochs in future works.

\noindent \textbf{Suspected Disease Obfuscation.}
EAR~\cite{park2024visual} uses DINO~\cite{caron2021emerging} attention maps to generate mask candidates; however, since DINO is not trained on biomedical data, it may produce inaccurate saliency predictions. To address this, we replace DINO with our CLAP method. We compare results using DINO~\cite{caron2021emerging} and BiomedCLIP~\cite{zhang2023biomedclip} within EAR to set mask candidates. Specifically, we use two BiomedCLIP~\cite{zhang2023biomedclip} configurations: (1) positive language prompting (PLP) alone and (2) CLAP, which combines both positive and negative prompts. Example prompts are provided in Table~\ref{table:prompts}.

\noindent \textbf{Evaluation Metric.}
We perform image-level detection on the BMAD dataset~\cite{bao2024bmad} and evaluate using AUROC. Anomaly scores are based on U-Net reconstruction error, illustrated in Fig.\ref{fig:cpmethod} (b). The reconstruction error between $I$ and $\hat{I}$ is computed via MSGMS, represented as \eqref{eq:loss_msgms}, following EAR~\cite{park2024visual}.

\begin{equation}
    \begin{aligned}
        \textit{MSGMS}(I, \hat{I}) = \sum_{s=1}^{S} \left( 1 - \frac{2g(I^{s})g(\hat{I}^{s}) + c}{g(I^{s})^2 + g(\hat{I}^{s})^2 + c} \right)
    \end{aligned}
    \label{eq:loss_msgms}
\end{equation}

\begin{table}[t]
    \vspace{-0.3cm}
    \caption{Comparison of anomaly detection performance on the BMAD dataset~\cite{bao2024bmad}, where PLP denotes positive language prompting, and our proposed method is CLAP.}
    \vspace{0.2cm}
    \centering
    \setlength{\tabcolsep}{4pt}
    \resizebox{\columnwidth}{!}{
        \begin{tabular}{l||c|c|c|c|c|c||c}
            \toprule
                \textbf{Anatomy} & \textbf{Brain MRI} & \textbf{Liver CT} & \multicolumn{2}{c|}{\textbf{Retinal OCT}} & \textbf{Chest X-ray} & \textbf{Lymph node} & \multirow{2}{*}{\textbf{Average}} \\
                \cmidrule(lr){1-1}\cmidrule(lr){2-2}\cmidrule(lr){3-3}\cmidrule(lr){4-5}\cmidrule(lr){6-6}\cmidrule(lr){7-7}
                \textbf{Dataset} & \textbf{\footnotesize{BraTS2021}} & \textbf{\footnotesize{BTCV + LiTs}} & \textbf{\footnotesize{RESC}} & \textbf{\footnotesize{OCT2017}} & \textbf{\footnotesize{RSNA}} & \textbf{\footnotesize{CAMELYON16}} & \\
            \hline
            \hline
                EAR~\cite{park2024visual}  
                    & 77.37 & 72.51 & 86.42 & \textbf{97.46} & \textbf{71.69} & 63.39 & 78.21 \\
                PLP  
                    & 73.54 & \textbf{72.76} & 90.08 & 96.77 & 65.23 & 64.98 & 77.23 \\
                CLAP (ours)  
                    & \textbf{78.55} & 72.60 & \textbf{91.66} & 96.38 & 65.76 & \textbf{68.42} & \textbf{78.89} \\
            \bottomrule
        \end{tabular}
    }
    \label{tab:performance}
    \vspace{-0.4cm}
\end{table}

\subsection{Qualitative results}
Qualitative results are shown in Fig.~\ref{fig:overview} and Fig.~\ref{fig:comparison}. Fig.~\ref{fig:overview} presents attention maps obtained with individual positive and negative prompts, compared to our method. Using only the positive prompt produces strong false positives in $A_{\textit{positive}}$, while the negative prompt $A_{\textit{negative}}$ includes some false negatives but yields mainly true negatives. This combination helps suppress false positives in $A_{\textit{positive}}$, resulting in the final attention map $A_{\textit{CLAP}}$.

We further compare the attention results of DINO~\cite{caron2021emerging} and CLAP in Fig.~\ref{fig:comparison}. DINO tends to highlight regions with distinct features, which often leads to unintended attention on non-lesion areas. In contrast, CLAP, built on BiomedCLIP~\cite{zhang2023biomedclip}, leverages language prompts to focus on specific lesion regions within an image. This capability allows BiomedCLIP to generate precise attention maps for lesions by modulating the language prompts. Designed to minimize false positive attention, CLAP demonstrates relatively accurate lesion localization compared to DINO, which aids in the accurate obfuscation of suspected defect regions before the image is input into the U-Net.

\subsection{Quantitative results}
To quantitatively assess the UAD performance, we report AUROC values in Table~\ref{tab:performance}. The primary objective of this study is to enhance the true positive detection capability of BiomedCLIP~\cite{zhang2023biomedclip} in biomedical anomaly detection tasks. 

Accordingly, we compare the attention performance of the non-biomedical model DINO~\cite{caron2021emerging} with that of BiomedCLIP using PLP, positive language prompting, and our CLAP method. The performance of basic DINO, marked with EAR, and PLP is almost the same. This means that DINO attention can catch the lesion to some extent without biomedical knowledge. At the same time, PLP has biomedical knowledge but contains false attention alarms. CLAP not only improves performance over PLP by attenuating false attention but also achieves better overall performance compared to EAR. Notably, CLAP performs particularly well on subsets of images with small, irregular patterns, such as those in `RESC' and `CAMELYON16'.

\section{Conclusion}
In this work, we introduced a novel approach, \textit{\textbf{C}ontrastive \textbf{LA}nguage \textbf{P}rompting (\textbf{CLAP})}, aimed at reducing false positives in medical anomaly detection using VLMs like BiomedCLIP~\cite{zhang2023biomedclip}. By utilizing both positive and negative prompts, our method enhances the identification of lesion regions while suppressing attention on normal areas, addressing a key limitation in previous models that only leveraged positive prompts. Extensive experiments with the BMAD dataset demonstrate that CLAP improves anomaly detection accuracy across various medical image types, outperforming both DINO~\cite{caron2021emerging} and single-prompt methods. Furthermore, we integrated CLAP with a reconstruction-by-inpainting U-Net approach, enhancing its diagnostic utility. Moving forward, we aim to refine this technique by automating language prompt generation to further support real-world clinical applications, reducing manual intervention and enhancing the scalability of this method in diverse medical scenarios.

\section{Acknowledgments}
\label{sec:acknowledgments}

This work was supported by SK Planet Co., Ltd., Korea. 

\bibliographystyle{IEEEbib}
\bibliography{refs}

\end{document}